\documentclass[letterpaper,journal]{IEEEtran}
\usepackage[T1]{fontenc}
\usepackage[utf8]{inputenc}
\usepackage{amsmath,amsfonts}
\usepackage{amssymb}
\usepackage{amsthm}
\usepackage{array}
\usepackage{booktabs}
\usepackage{textcomp}
\usepackage{stfloats}
\usepackage{url}
\usepackage{graphicx}
\usepackage[table]{xcolor}
\usepackage{tabularx}
\usepackage{placeins}
\usepackage{calc}
\usepackage{cite}
\usepackage[bookmarks=true,bookmarksnumbered=true,bookmarksopen=true,bookmarksopenlevel=1,
 breaklinks=false,pdfborder={0 0 1},backref=false,colorlinks=false]{hyperref}
\hypersetup{pdftitle={FIRE-Drive: Feasibility-Informed Risk Evaluation for End-to-End Diffusion Planning},
 pdfauthor={Qifeng Xie, Kai-Yuan Feng, Wenkang Wan, Yue Wu, Maoguo Gong, Kai Sheng},pdfsubject={Preprint: End-to-end autonomous driving and risk-aware trajectory planning},pdfborderstyle=,pdfpagelayout=TwoColumnRight,pdfnewwindow=true,pdfstartview=XYZ,plainpages=false}
\newcolumntype{Y}{>{\centering\arraybackslash}X}

\theoremstyle{plain}

\begin{document}
\title{FIRE-Drive: Feasibility-Informed Risk Evaluation for End-to-End Diffusion Planning}
\author{Qifeng~Xie, Kai-Yuan~Feng,~\IEEEmembership{Member,~IEEE,}
Wenkang~Wan, Yue~Wu,~\IEEEmembership{Senior~Member,~IEEE,}
Maoguo~Gong,~\IEEEmembership{Fellow,~IEEE,} and Kai~Sheng%
\thanks{Corresponding author: Kai-Yuan Feng.}%
\thanks{Qifeng Xie is with the School of Electronic Engineering, Xidian University, and also with the Key Laboratory of Collaborative Intelligence Systems, Ministry of Education, Xidian University, Xi'an, Shaanxi Province 710071, China (e-mail: 24021110872@stu.xidian.edu.cn).}%
\thanks{Kai-Yuan Feng and Kai Sheng are with the Faculty of Infor-X, Xidian University, and also with the Key Laboratory of Collaborative Intelligence Systems, Ministry of Education, Xidian University, Xi'an, Shaanxi Province 710071, China (e-mail: fkylwl@gmail.com; kaisheng@xidian.edu.cn).}%
\thanks{Wenkang Wan is with the Key Laboratory of Collaborative Intelligence Systems, Ministry of Education, Xidian University, Xi'an, Shaanxi Province 710071, China (e-mail: wanwk@stu.xidian.edu.cn).}%
\thanks{Yue Wu is with the School of Computer Science and Technology, Xidian University, and also with the Key Laboratory of Collaborative Intelligence Systems, Ministry of Education, Xidian University, Xi'an, Shaanxi Province 710071, China (e-mail: ywu@xidian.edu.cn).}%
\thanks{Maoguo Gong is with the Key Laboratory of Collaborative Intelligence Systems, Ministry of Education, Xidian University, Xi'an, Shaanxi Province 710071, China, and also with the College of Artificial Intelligence, Inner Mongolia Normal University, Hohhot 010028, China (e-mail: gong@ieee.org).}}

\markboth{Preprint}{FIRE-Drive -- Preprint}
\maketitle
\begin{abstract}
Diffusion-based end-to-end autonomous driving models have shown strong potential in generating multimodal planning trajectories. However, their performance still depends heavily on trajectory initialization, mode evaluation, and supervision quality. Weakly constrained noisy initialization may introduce infeasible proposals and increase the burden of denoising. Moreover, selecting the optimal mode remains challenging when candidate trajectories are geometrically similar but differ in safety, interaction risk, and progress. To address these issues, we propose FIRE-Drive, a feasibility-informed risk evaluation framework for end-to-end autonomous driving. FIRE-Drive improves diffusion-based planning through dynamic feasible anchor initialization, candidate-specific bird's-eye-view risk evidence, and compact high-quality candidate supervision with lightweight refinement. This design aligns the initial support, risk-aware scoring, and multimodal supervision of trajectory modes under a unified planning objective. It enables the diffusion decoder to focus on refining plausible trajectories while the scorer distinguishes modes using local risk evidence. Experiments on NAVSIM v1 show that FIRE-Drive achieves a Predictive Driver Model Score of 91.5 with a ResNet34 backbone. Additional closed-loop evaluation on 220 Bench2Drive routes yields a Driving Score of 71.43 and a Success Rate of 42.73\%. Component ablations and a scorer comparison under matched candidate inputs support the contributions of feasible initialization, auxiliary supervision, and risk-informed selection. Our code is available at \url{https://github.com/Latte00/FIRE-drive}.
\end{abstract}


\IEEEpeerreviewmaketitle{}

\section{Introduction}

\IEEEPARstart{E}{nd}-to-end (E2E) autonomous driving has recently emerged as a promising paradigm for unifying perception, prediction, and planning within a single learnable framework~\cite{Hu_2023_CVPR_UniAD,Jiang_2023_ICCV_VAD}. Compared with modular pipelines, E2E planners can reduce hand-crafted interface mismatches and can directly optimize the final driving behavior from scene observations. A central difficulty, however, is that human driving futures are inherently multimodal: the same traffic scene may admit multiple plausible maneuvers, interaction patterns, and risk-efficiency trade-offs~\cite{Nayakanti_2023_ICRA_Wayformer}. Therefore, an effective E2E planner should not only generate diverse future trajectories, but also ensure that the generated candidates are feasible, distinguishable, and properly supervised.

\begin{figure}[t]
	\centering
	\includegraphics[width=\columnwidth]{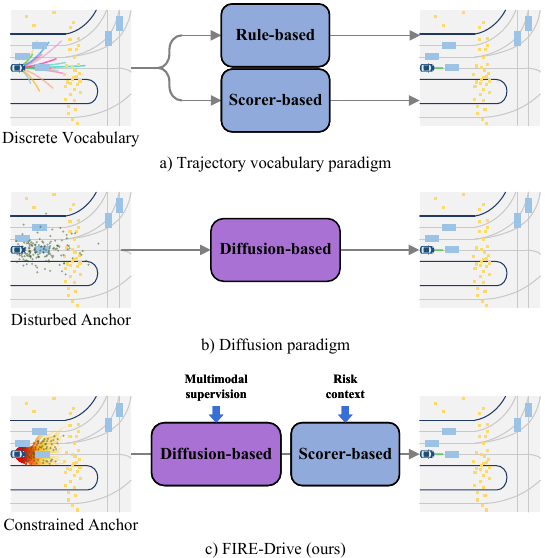}
	\caption{Comparison of representative E2E planning paradigms. Dense vocabulary methods first construct a large candidate table and then select a trajectory by rules or scoring; conventional diffusion planners decode from noisy or perturbed anchors; FIRE-Drive introduces feasibility-informed anchor initialization and risk-aware evaluation before final selection.}\label{fig:intro_anchor_comparison}
\end{figure}

A common planning strategy constructs a trajectory vocabulary or proposal pool and then selects a plan using learned scores~\cite{Li_2024_arXiv_HydraMDP,Li_2025_arXiv_HydraMDPpp}, as illustrated in Fig.~\ref{fig:intro_anchor_comparison}. This separates candidate coverage from decision making, but also creates two coupled requirements: the pool must contain useful alternatives, and the scorer must identify them. A selector cannot recover a feasible maneuver absent from its candidate set, while geometrically similar candidates may require different decisions because of their local collision exposure or road-boundary proximity. Candidate generation and evaluation should therefore be designed together.

Diffusion-based planners offer continuous multimodal generation through denoising~\cite{Liao_2025_CVPR_DiffusionDrive,Liu_2025_arXiv_BridgeDrive}, but this flexibility leaves three challenges. First, fixed anchors or generic perturbations need not respect the current drivable region, obstacle layout, or ego-motion reachability, leaving the decoder to correct scene-inconsistent starting points. Physical priors help structure trajectory modeling~\cite{Zheng_2026_CVPR_ResAD}, yet scene-dependent admissibility also matters. Second, geometric plausibility alone does not determine driving quality: similar trajectories can differ in safety and progress, requiring candidate-specific risk evidence for selection. Third, supervision from a single logged future provides limited guidance for alternative maneuvers. Additional references are useful only when their diversity is accompanied by scene compatibility and adequate driving quality.

The distinction between motion plausibility and scene feasibility is particularly important at initialization. A trajectory can be smooth and consistent with the current vehicle state while entering occupied space or extending beyond a drivable boundary. Conversely, an unoccupied road region may be difficult to reach under the current speed and steering limits. A useful initial candidate should account for both restrictions. This motivates a scene-dependent admissible region that combines semantic drivability with ego-motion reachability, rather than relying on a motion prior alone to locate the starting hypotheses.

Selection presents a related but different problem. Consider two candidates that follow nearly the same lane but differ slightly in lateral clearance or longitudinal progress. Their trajectory embeddings may be similar, yet one can pass closer to a vehicle or road boundary. The relevant evidence is therefore not just what objects exist in the scene, but where each candidate travels relative to them. Candidate-conditioned risk context makes these spatial relationships available to the scorer, while explicit risk supervision encourages attention to the regions that explain potential infractions.

The interplay between training data and learning strategies is central to E2E driving~\cite{Xu_2026_TITS_TrainingSurvey}. Additional supervision must therefore preserve the distinction between possible and useful behavior. The logged future provides a demonstrated response, but does not enumerate all acceptable maneuvers. Geometric variation can supply alternatives, although larger variation does not necessarily imply better supervision. We therefore treat reference construction as a quality-control step: behavioral alternatives are screened against scene geometry and evaluator scores before they become training targets. Caching this compact reference set separates its construction cost from repeated training updates.

We propose FIRE-Drive to address these challenges through coordinated initialization, evaluation, and supervision. Scene-adaptive anchor priors are constrained by semantic drivability and ego-motion reachability before denoising. A candidate-conditioned bird's-eye-view (BEV) risk context supplies local evidence for trajectory selection. Compact, quality-filtered auxiliary references supervise secondary modes and support lightweight refinement; their trajectories and evaluator scores are cached offline for reuse. Together, these components improve the starting hypotheses, the training targets, and the evidence used to choose the final plan.

The contributions of this paper are summarized as follows:
\begin{itemize}
    \item We introduce a dynamic feasible anchor strategy that replaces weakly constrained seeds with scene-consistent trajectory priors constructed under semantic drivability and motion-reachability constraints.
    \item We propose a mode risk context module that combines candidate-conditioned BEV evidence, semantic risk bias, and infraction supervision to support safety-aware trajectory selection.
    \item We formulate a compact procedure for high-quality candidate supervision, combining behavior-based reference generation, geometry and quality filtering, and offline caching to supervise secondary modes without repeated reference evaluation.
    \item FIRE-Drive achieves 91.5 PDMS on NAVSIM v1 and 71.43 DS with 42.73\% SR over 220 Bench2Drive routes. Component and matched-input scorer ablations evaluate the contributions of the proposed designs.
\end{itemize}

\section{Related Work}

\subsection{E2E Autonomous Driving}

E2E driving has evolved from direct image-to-control imitation~\cite{Bojarski_2016_arXiv_PilotNet,Codevilla_2018_ICRA_CIL,LeMero_2022_TITS_ImitationSurvey} toward structured systems that learn perception and planning jointly~\cite{Chen_2024_TPAMI_E2ESurvey,Hu_2023_CVPR_UniAD}. Closed-loop simulation exposes the consequences of prediction errors beyond individual frames~\cite{Dosovitskiy_2017_CoRL_CARLA}, motivating trajectory-guided control and iterative decoding~\cite{Wu_2022_NeurIPS_TCP,Jia_2023_CVPR_ThinkTwice}. Richer sensor fusion and temporal reasoning improve the scene evidence available to the planner~\cite{Chitta_2022_TPAMI_TransFuser,Shao_2023_CVPR_ReasonNet}, while vectorized and sparse representations reduce dependence on dense scene processing~\cite{Jiang_2023_ICCV_VAD,Jiang_2026_ICLR_VADv2,Sun_2025_ICRA_SparseDrive}.

Another direction strengthens the relationship between predicted interactions and ego decisions through joint optimization, hierarchical prediction, or game-theoretic reasoning~\cite{Huang_2024_TNNLS_DIPP,Liu_2025_TPAMI_HPP,Huang_2023_ICCV_GameFormer}. Unified transformer architectures further facilitate information exchange across driving tasks~\cite{Jia_2025_ICLR_DriveTransformer}. However, richer representations and tighter task coupling alone do not ensure that every generated mode is feasible or receives useful supervision. Studies of PDM and PlanTF also show that good open-loop fitting does not directly translate into reliable closed-loop behavior~\cite{Dauner_2023_CoRL_PDM,Cheng_2024_ICRA_PlanTF}. This motivates distinguishing scene representation from the use of that representation in planning. Even with a shared perception backbone, different initial candidate supports, training targets, and selection criteria can lead to different driving decisions. FIRE-Drive addresses these proposal-level choices within a diffusion planner, complementing advances in scene encoding and interaction modeling.

\subsection{Diffusion Models for Trajectory Prediction and Planning}

Denoising-based generation~\cite{Ho_2020_NeurIPS_DDPM,Song_2021_ICLR_DDIM} provides a flexible representation of multimodal trajectories and actions~\cite{Janner_2022_ICML_Diffuser,Chi_2023_RSS_DiffusionPolicy}. Its application to driving spans multi-agent forecasting and joint prediction--planning~\cite{Jiang_2023_CVPR_MotionDiffuser,Zheng_2025_ICLR_DiffusionPlanner}. Truncated diffusion, goal-driven flow matching, and diffusion bridges adapt this formulation to E2E trajectory generation~\cite{Liao_2025_CVPR_DiffusionDrive,Xing_2025_CVPR_GoalFlow,Liu_2025_arXiv_BridgeDrive}. Their flexibility makes feasibility control important: representing multiple futures does not by itself make those futures admissible in the current scene.

Physical structure enters existing methods in different forms: ResAD predicts normalized residuals around a deterministic inertial reference~\cite{Zheng_2026_CVPR_ResAD}, and WAKE integrates vehicle kinematics into trajectory prediction~\cite{Wang_2025_TPAMI_WAKE}. Diffusion-Phy adopts a multi-stage physical-prior pipeline involving dynamic-state prediction, neural compensation, and latent-state encoding to construct adaptive anchors~\cite{Hu_2026_PR_DiffusionPhy}. FIRE-Drive adopts a simpler initialization design, emphasizing the spatial admissibility of the denoising seeds through explicit semantic--reachability constraints, while leaving detailed trajectory shaping to the decoder and refiner. Specifically, the intersection of semantic and reachable regions restricts sampled anchor locations, beyond using physical features to condition their distribution. Other approaches enforce constraints during flow generation~\cite{Liu_2026_CVPR_GuideFlow} or use reward-based evolutionary search~\cite{Yang_2024_CVPR_DiffusionES}; our design places the spatial restriction before decoding.

Reward-based learning provides another route to improving candidate quality. DIVER combines multiple references and one-to-one matching with GRPO rewards for diversity and driving quality~\cite{Song_2025_arXiv_DIVER}; DiffusionDriveV2 uses anchor-aware GRPO and a trajectory-conditioned selector~\cite{Zou_2025_arXiv_DiffusionDriveV2}. Their policy updates involve trajectory sampling and reward evaluation, adding computationally demanding online evaluation to training. FIRE-Drive uses offline-cached, quality-filtered references as reusable supervised targets for secondary modes. This avoids repeated reward evaluation for auxiliary reference supervision, while retaining online infraction evidence for risk-context training.

\subsection{Mode Evaluation and Multimodal Supervision}

Learning to rank proposals is increasingly important as candidate generation improves. Scorer distillation and dedicated trajectory representations learn to distinguish candidate quality~\cite{Li_2024_arXiv_HydraMDP,Li_2025_arXiv_HydraMDPpp,Kirby_2026_arXiv_DrivingOnRegisters,Li_2025_arXiv_GTRS}, while recent proposal-centric systems further emphasize the interaction between generation and selection~\cite{Guo_2025_arXiv_iPad,Sun_2026_arXiv_SparseDriveV2,Yao_2025_arXiv_DriveSuprim}. Future-scene modeling and safety-oriented value estimation provide additional ways to assess downstream outcomes~\cite{Wang_2024_CVPR_DriveWM,Gao_2026_TSMC_SoAD}. Explicit risk fields and risk-aware optimization connect spatial hazard assessment with trajectory planning~\cite{Tian_2025_TITS_RiskField,Lin_2025_TITS_Contingency}, while risk-aware prediction incorporates safety-critical interactions into future motion estimation~\cite{Wang_2024_arXiv_RiskAwareTP}. FIRE-Drive brings spatial risk evidence into learned candidate scoring: candidate-conditioned BEV context, semantic risk bias, and infraction supervision guide the scorer toward hazards relevant to each trajectory, beyond aggregate score regression alone. The matched-input scorer ablation supports the benefit of this risk evidence independently of changes in candidate quality.

A separate issue is what supervises the alternatives being scored. Learned latent anchors and multimodal aggregation improve the representation of possible futures~\cite{Varadarajan_2022_ICRA_MultiPathPP}, and mode assignment directs supervision to predictions compatible with the observed future~\cite{Shi_2022_NeurIPS_MTR,Zhou_2023_CVPR_QCNet}. However, a single observation cannot specify the quality of every alternative maneuver. Multi-reference supervision broadens the available targets~\cite{Liu_2026_TPAMI_R2SE}, but reference diversity alone does not establish scene feasibility or driving utility. FIRE-Drive explicitly screens auxiliary references by geometry and evaluator quality, then uses candidate matching and bounded refinement to supervise secondary modes. This connects broader supervision to a compact set of evaluated alternatives.

\begin{figure*}[t]
\centering
\includegraphics[width=\textwidth]{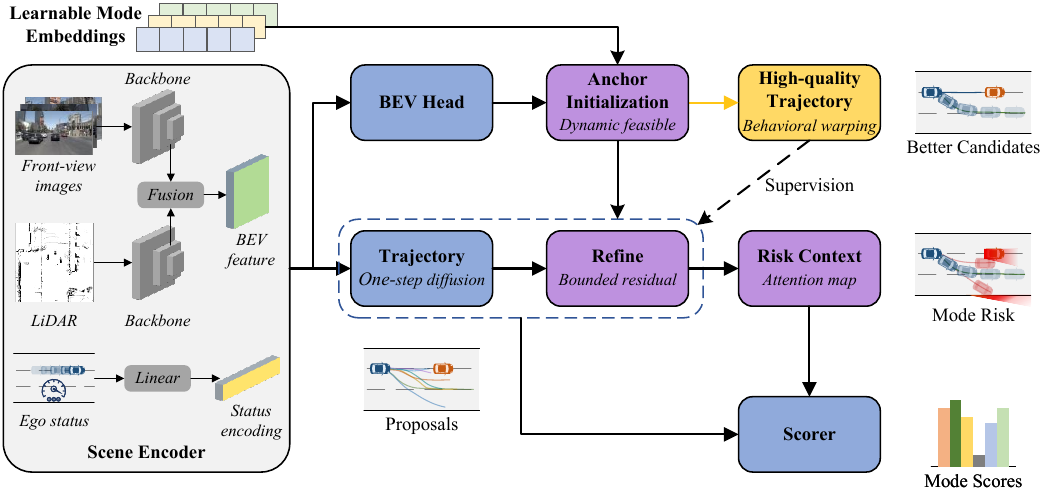}
\caption{Overall framework of the proposed method. FIRE-Drive intervenes in diffusion planning through feasibility-informed anchor initialization, mode-conditioned risk evaluation, and compact candidate-guided refinement.}\label{fig:overall_framework}
\end{figure*}

\section{Methodology}

\subsection{Overall Framework}

FIRE-Drive takes a stitched image from the front-left, front, and front-right cameras, LiDAR measurements, and ego status as input. The perception backbone encodes these observations into scene features $\chi$, including BEV features and semantic predictions. As shown in Fig.~\ref{fig:overall_framework}, scene-conditioned anchors initialize the diffusion decoder, which predicts $K$ trajectories $Y_k$ with $T$ waypoints containing position and heading. A bounded refiner produces $\hat Y_k$, and a risk-aware scorer selects the final plan:
\begin{equation}
Y^*=\hat Y_{k^*},\qquad k^*=\arg\max_k S(\hat Y_k,\chi).
\end{equation}
When refinement is disabled, $\hat Y_k=Y_k$. During training, offline auxiliary references supervise secondary modes, while localized infraction evidence supervises the scorer's risk attention.

Initialization defines the decoder's starting hypotheses, candidate supervision supplies additional learning targets, and risk context informs selection among the refined outputs. Scene features are shared across modes, while anchor embeddings and trajectory queries preserve mode-specific representations.

\subsection{Dynamic Feasible Anchor Initialization}

\begin{figure}[t]
	\centering
	\includegraphics[width=0.92\linewidth]{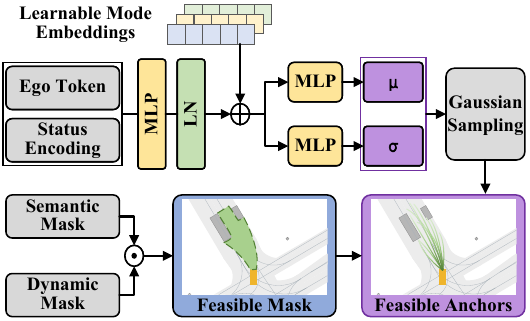}
	\caption{Illustration of dynamic feasible anchor initialization. Semantic drivability and ego-motion reachability jointly define the admissible region for anchor initialization.}\label{fig:dynamic_feasible_anchor}
\end{figure}

Fixed anchors and unconstrained noise can place initial hypotheses outside the locally admissible driving region~\cite{Liao_2025_CVPR_DiffusionDrive,Liu_2025_arXiv_BridgeDrive}. We instead combine a scene-conditioned Gaussian prior with an explicit feasible mask, as illustrated in Fig.~\ref{fig:dynamic_feasible_anchor}.

The ego-status encoding and ego query are fused into a shared context, which is combined with a learnable embedding for each mode to obtain $z_k$. Prediction heads produce a residual mean $\Delta\mu_k$ and a positive, bounded standard deviation $\sigma_k$. A motion prior $\mu_k^{\mathrm{phy}}$ estimated from ego velocity and acceleration provides the reference mean~\cite{Zheng_2026_CVPR_ResAD}:
\begin{equation}
\hat A_k\sim\mathcal N\!\left(\mu_k^{\mathrm{phy}}+\alpha\Delta\mu_k,
\operatorname{diag}(\sigma_k^2)\right),
\end{equation}
where $\alpha$ scales the learned residual. Softplus and clipping keep the predicted standard deviations positive and bounded.

The motion prior captures continuation of the current ego state, while the learned residual adapts this reference to the encoded scene. Separate mode embeddings allow the prediction heads to produce different means and uncertainty scales from the same observation. The Gaussian samples provide alternative starting hypotheses, but their spatial support is further shaped by the feasible mask. In this way, the learned distribution specifies where candidates are likely to lie, and the mask specifies which sampled locations are admissible under the current scene estimate.

To incorporate scene geometry, the semantic BEV mask $M_{\mathrm{sem}}$ retains drivable regions and suppresses occupied space. A reachable mask $M_{\mathrm{reach}}$ is obtained by propagating a simplified bicycle model under admissible acceleration and steering sequences and projecting reachable positions onto BEV. Their intersection defines the admissible region:
\begin{equation}
M_{\mathrm{fea}}=M_{\mathrm{sem}}\odot M_{\mathrm{reach}}.
\end{equation}

The two masks provide complementary information. Semantic drivability alone can retain road regions that are difficult to reach from the current vehicle state. Reachability alone can include locations occupied by another agent or outside the road. Their intersection incorporates both restrictions into the initialization, allowing the decoder to begin from hypotheses informed by scene layout as well as ego motion. The mask is reconstructed from the current observation at each planning cycle.
Samples are then projected into this region over a configurable constraint horizon $T_c$:
\begin{equation}
A_k^{(t)}=
\begin{cases}
\Pi_{M_{\mathrm{fea}}}(\hat A_k^{(t)}),&t\leq T_c,\\
\hat A_k^{(t)},&t>T_c.
\end{cases}
\end{equation}
Here $\Pi_{M_{\mathrm{fea}}}$ maps inadmissible samples to the estimated feasible region. This constrains initialization; subsequent decoding and selection determine final-trajectory feasibility.

The anchors enter the decoder directly, without an additional forward-noising step. The decoder attends to BEV and surrounding-agent features and performs a single residual denoising update. An anchor loss $\mathcal L_{\mathrm{anc}}$ minimizes the mean negative log-likelihood of matched target anchors under the predicted Gaussian priors.

\subsection{Mode Risk Context}

\begin{figure*}[t]
\centering
\includegraphics[width=\textwidth]{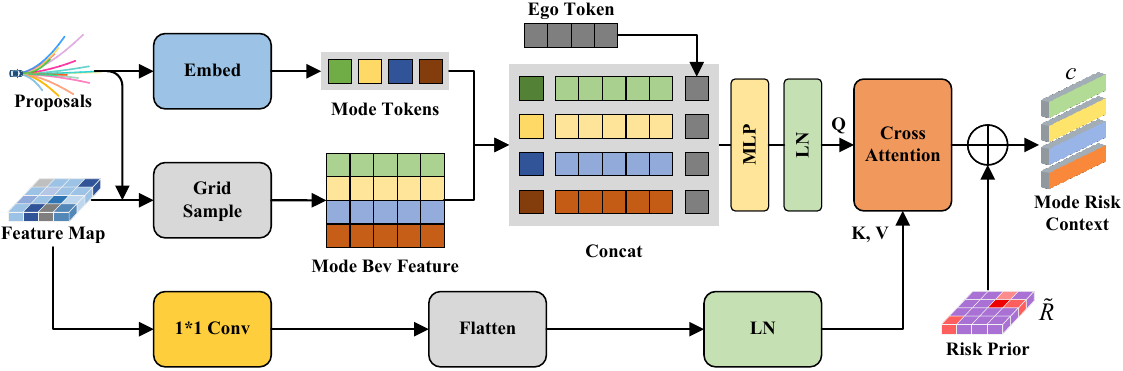}
\caption{Illustration of the mode risk context module. Candidate-specific trajectory queries extract local BEV risk evidence, which is fused into the scoring branch to distinguish geometrically similar modes.}\label{fig:mode_risk_context}
\end{figure*}

Geometrically similar trajectories may pass through different obstacle or boundary regions. The mode risk context module in Fig.~\ref{fig:mode_risk_context} makes this local evidence available before trajectory scoring.

We construct a semantic BEV risk prior at position $P$:
\begin{equation}
R(P)=\lambda_{\mathrm{off}}\bigl(1-p_{\mathrm{drv}}(P)\bigr)
+\lambda_{\mathrm{obs}}p_{\mathrm{obs}}(P),
\end{equation}
where $p_{\mathrm{drv}}$ is the maximum of road and centerline probabilities, and $p_{\mathrm{obs}}$ aggregates obstacle occupancy, including pedestrians. Spatial-maximum normalization with a numerical stabilizer gives $\tilde R$. Target BEV labels supply the prior during training, while predicted semantics supply it at inference.

The risk prior is shared across candidate modes and describes spatial hazards in the scene. It is not itself a trajectory score: the same obstacle may be relevant to one candidate and distant from another. Candidate conditioning is introduced through the trajectory query, which carries sampled features along the evaluated path together with the mode representation and ego state. This separates the scene-level location of potential hazards from their relevance to a particular trajectory.

For each decoded trajectory $Y_k$, we fuse sampled BEV features, its mode token, and ego encoding into a query $q_k$. Let $b_j\in\mathbb R^d$ be the projected BEV token at spatial position $j$. Risk-biased attention aggregates the candidate context as
\begin{equation}
\begin{aligned}
a_{k,j}&=\operatorname{softmax}_{j}\!\left(
\frac{q_k^\top b_j/\sqrt d+\beta\tilde R_j}{\tau_{\mathrm{attn}}}\right),\\
c_k&=\sum_j a_{k,j}b_j,
\end{aligned}
\end{equation}
where $\beta$ controls risk bias and $\tau_{\mathrm{attn}}$ is the attention temperature. The implementation centers the risk map spatially before adding it; this constant shift leaves the softmax unchanged. A learned pre-fusion module combines $c_k$ with the trajectory token before the multi-component PDM scorer decodes its scores.

The content term in attention measures compatibility between the candidate query and a BEV location, while the risk term increases the salience of potentially hazardous regions. Their combination enables the context to represent nearby conflicts even when two candidates have similar overall shapes. Because the context is fused before score decoding, the scorer can learn how this evidence relates to each predicted quality component rather than receiving only a fixed penalty after prediction. The score remains a learned assessment of the complete candidate.

The auxiliary loss $\mathcal L_{\mathrm{attn}}$ aligns attention with online collision and off-road evidence. Event locations are projected onto BEV and represented by Gaussian targets, with exponentially decaying temporal weights bounded below by a minimum weight. After normalization, the targets supervise predicted attention through target-to-prediction KL divergence. Gaussian width, decay, minimum weight, and loss strength are hyperparameters. This supplies localized risk supervision in addition to trajectory-score targets.

Scalar quality labels indicate whether a candidate is desirable, but do not identify which spatial region accounts for a collision or boundary violation. The event-based attention targets provide this localization. Gaussian spreading gives nearby BEV cells a graded training signal, and temporal weighting emphasizes earlier events while retaining a contribution from later ones. Risk bias and attention supervision therefore serve different roles: one guides feature aggregation, while the other trains that aggregation using observed infractions.

\subsection{High-Quality Candidate Supervision and Refinement}

\textbf{Reference construction.} Left-overtake, right-overtake, and acceleration warping operators construct auxiliary trajectories from the logged future. Up to five references and their evaluator scores are cached per scene. With $\mathcal C$ denoting the non-padding pool and $G$ the geometry-validity indicator, the effective set is
\begin{equation}
\mathcal C^+=\{C_m\in\mathcal C:G(C_m)=1,\ Q(C_m)>Q(Y^{\mathrm{gt}})\}.
\end{equation}
Geometry checks limit the absolute wrapped heading change between adjacent waypoints to 1.5 rad and allow at most one sign reversal in lateral increments; the separate heading-reversal count filter is disabled. On NAVSIM, $Q$ is the cached PDM evaluator score, distinct from the learned scorer $S$. References and scores are prepared offline and reused during training; neither is needed at inference.

The score comparison is scene-specific: each reference is compared with the logged trajectory under the same cached evaluator. Consequently, the number of effective references varies across scenes even though the cache has a fixed maximum capacity. A scene with no qualifying reference still provides the ordinary GT supervision. The additional targets extend supervision only where the cached pool offers an eligible alternative, rather than forcing every scene to contribute the same number of secondary targets.

\textbf{Candidate matching.} Excluding the GT-assigned mode, we sort candidate--mode pairs by mean waypoint distance in the $xy$ plane and greedily accept pairs whose candidate and mode have not been used. The loss $\mathcal L_{\mathrm{cand}}$ averages distances over accepted pairs. Empty sets skip this auxiliary loss while retaining the primary training objectives. Candidate matching remains active without refinement and then supervises decoder outputs directly.

\textbf{Bounded refinement.} A lightweight refiner uses the mode token and ego encoding to predict residual corrections, clipped componentwise to $\pm1.0$ m in $x$ and $y$ and $\pm0.1$ rad in heading per refinement step. Its loss $\mathcal L_{\mathrm{ref}}$ combines the minimum distance from a refined mode to GT with the mean nearest-reference distance for the remaining modes. Multiple modes may share a nearest reference. Samples without effective candidates contribute zero to the reference term, and the loss weights are normalized over the terms available in the minibatch. 

\begin{table*}[!tp]
\centering
\caption{Evaluation on NAVSIM v1 with official NAVSIM metrics. The gray human-reference row is retained for context. Within each separated block, the best and second-best results are bolded and underlined, respectively.}\label{tab:navsim_v1_results}
\footnotesize
\setlength{\tabcolsep}{5pt}
\renewcommand{\arraystretch}{1.08}
\begin{tabularx}{\linewidth}{l l Y Y Y Y Y Y}
\toprule
Method & Backbone & NC$\uparrow$ & DAC$\uparrow$ & EP$\uparrow$ & TTC$\uparrow$ & C$\uparrow$ & PDMS$\uparrow$ \\
\midrule
Transfuser~\cite{Chitta_2022_TPAMI_TransFuser} (2022) & ResNet34 & 97.7 & 92.8 & 79.2 & 92.8 & \textbf{100} & 84.0 \\
UniAD~\cite{Hu_2023_CVPR_UniAD} (2023) & ResNet34 & 97.8 & 91.9 & 78.8 & 92.9 & \textbf{100} & 83.4 \\
Hydra-MDP~\cite{Li_2024_arXiv_HydraMDP} (2024) & ResNet34 & 98.3 & 96.0 & 78.7 & 94.6 & \textbf{100} & 86.5 \\
Hydra-MDP++~\cite{Li_2025_arXiv_HydraMDPpp} (2025) & ResNet34 & 97.6 & 96.0 & 80.4 & 93.1 & \textbf{100} & 86.6 \\
GoalFlow~\cite{Xing_2025_CVPR_GoalFlow} (2025) & ResNet34 & 98.3 & 93.8 & 79.8 & 94.3 & \textbf{100} & 85.7 \\
DiffusionDrive~\cite{Liao_2025_CVPR_DiffusionDrive} (2025) & ResNet34 & 98.2 & 96.2 & 82.2 & 94.7 & \textbf{100} & 88.1 \\
DiffusionDriveV2~\cite{Zou_2025_arXiv_DiffusionDriveV2} (2025) & ResNet34 & 98.3 & \textbf{97.9} & \underline{87.5} & 94.8 & \underline{99.9} & \underline{91.2} \\
DIVER~\cite{Song_2025_arXiv_DIVER} (2025) & ResNet34 & \underline{98.5} & 96.5 & 82.6 & \underline{94.9} & \textbf{100} & 88.3 \\
ResAD~\cite{Zheng_2026_CVPR_ResAD} (2026) & ResNet34 & 98.0 & \underline{97.5} & 83.3 & 94.1 & \textbf{100} & 88.8 \\
DriveSuprim~\cite{Yao_2025_arXiv_DriveSuprim} (2026) & ResNet34 & 97.8 & 97.3 & 86.7 & 93.6 & \textbf{100} & 89.9 \\
\rowcolor{gray!12}
FIRE-Drive (Ours) & ResNet34 & \textbf{98.9} & \underline{97.5} & \textbf{87.6} & \textbf{95.8} & \textbf{100} & \textbf{91.5} \\
\midrule
Hydra-MDP~\cite{Li_2024_arXiv_HydraMDP} (2024) & V2\mbox{-}99 & 98.4 & \underline{97.8} & \underline{86.5} & 93.9 & \textbf{100} & 90.3 \\
Hydra-MDP~\cite{Li_2024_arXiv_HydraMDP} (2024) & ViT-L & 98.4 & 97.7 & 85.0 & 94.5 & \textbf{100} & 89.9 \\
ResAD~\cite{Zheng_2026_CVPR_ResAD} (2026) & V2\mbox{-}99 & \textbf{98.9} & \underline{97.8} & \textbf{87.0} & 94.9 & \textbf{100} & \underline{90.6} \\
Hydra-MDP++~\cite{Li_2025_arXiv_HydraMDPpp} (2025) & V2\mbox{-}99 & \underline{98.6} & \textbf{98.6} & 85.7 & \underline{95.1} & \textbf{100} & \textbf{91.0} \\
VADv2~\cite{Jiang_2026_ICLR_VADv2} (2026) & BEVFormer & 98.3 & 97.4 & 82.3 & \textbf{95.7} & \textbf{100} & 89.3 \\
\midrule
\textcolor{gray}{Human Agent} & \textcolor{gray}{\textemdash{}} & \textcolor{gray}{100} & \textcolor{gray}{100} & \textcolor{gray}{87.5} & \textcolor{gray}{100} & \textcolor{gray}{99.9} & \textcolor{gray}{94.8} \\
\bottomrule
\end{tabularx}
\end{table*}

\begin{figure*}[!t]
	\centering
	\includegraphics[width=\textwidth]{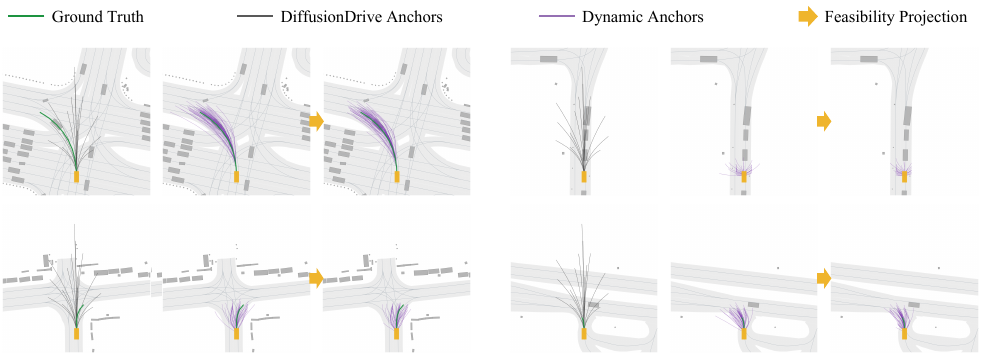}
	\caption{Qualitative comparison of anchor initialization in four driving scenes. Each group shows DiffusionDrive anchors (gray), dynamic anchors before projection (purple), and dynamic anchors after projection (purple), from left to right. Green denotes the ground-truth trajectory, and yellow arrows indicate feasibility projection. The comparison illustrates local changes in the constrained anchor prefix rather than a full-horizon feasibility guarantee.}\label{fig:anchor_initialization_comparison}
\end{figure*}

\subsection{Training Objective}

The primary perception and planning loss $\mathcal L_{\mathrm{base}}$ is augmented by the anchor, attention, candidate-matching, and refinement losses:
\begin{equation}
\begin{aligned}
\mathcal L_{\mathrm{total}}={}&\mathcal L_{\mathrm{base}}
+\lambda_{\mathrm{anc}}\mathcal L_{\mathrm{anc}}
+\lambda_{\mathrm{attn}}\mathcal L_{\mathrm{attn}}\\
&+\eta_{\mathrm{cand}}\mathcal L_{\mathrm{cand}}
+\lambda_{\mathrm{ref}}\mathcal L_{\mathrm{ref}}.
\end{aligned}
\end{equation}
The coefficients balance the supervised objectives. Candidate matching is applied at supervised decoder outputs, using the refined final output when refinement is enabled.

We use a candidate-matching weight of 1.0 and a refinement auxiliary weight of 0.1 within the decoder loss. The GT and auxiliary-reference terms in refinement have equal relative weights of 1.0, normalized over the available terms. The attention KL loss is weighted by 0.1 when added to the overall objective. These are module-level weights; the decoder terms also inherit the base decoder-loss scaling.

\section{Experiments}

\subsection{Experimental Setup}

\textit{1. NAVSIM.}~\cite{Dauner_2024_NeurIPS_NAVSIM} We use NAVSIM v1 as the main benchmark, evaluating quantitative comparisons and ablations on its navtest split. Following the official non-reactive simulation protocol, we report Non-at-fault collision (NC), drivable area compliance (DAC), ego progress (EP), time-to-collision (TTC), comfort (C), and their weighted aggregation, namely the Predictive Driver Model Score (PDMS), defined as

\begin{equation}
\mathrm{PDMS}=\mathrm{NC}\cdot \mathrm{DAC}\cdot \frac{5\,\mathrm{TTC}+2\,\mathrm{C}+5\,\mathrm{EP}}{12}.
\end{equation}

The equation uses per-scene component scores on a 0--1 scale; tables report dataset averages on a 0--100 scale. Detailed definitions of the individual metrics are provided in the NAVSIM benchmark paper~\cite{Dauner_2024_NeurIPS_NAVSIM}.

\textit{2. Bench2Drive.}~\cite{Jia_2024_NeurIPS_Bench2Drive} We evaluate closed-loop driving in CARLA 0.9.15 on 220 Bench2Drive routes. Driving Score (DS) averages the per-route product of route completion and an infraction penalty factor. Success Rate (SR) is the percentage of evaluated routes completed successfully without disqualifying infractions under the benchmark criteria.

\textit{3. Implementation details.} Unless otherwise specified, all experiments follow the main model configuration. To better isolate planning performance, we use a ResNet34 backbone~\cite{He_2016_CVPR_ResNet} instead of a larger perception backbone. The input format follows DiffusionDrive~\cite{Liao_2025_CVPR_DiffusionDrive}; the planner predicts 64 ego future modes, uses a 2-layer diffusion decoder, and applies one refinement step. The anchor constraint horizon is set to $T_c=3$ steps. Training is performed with AdamW using a learning rate of $2\times10^{-4}$, a batch size of 64, and mixed precision on an NVIDIA RTX 5090 GPU. The main model is evaluated using the checkpoint from epoch 40.

\begin{table}[!htbp]
\centering
\caption{Reported Bench2Drive closed-loop results. DS is Driving Score and SR is Success Rate. Best and second-best entries are bolded and underlined. Other methods are literature results; FIRE-Drive is evaluated on 220 routes.}
\label{tab:bench2drive_results}
\footnotesize
\setlength{\tabcolsep}{4pt}
\renewcommand{\arraystretch}{1.08}
\begin{tabularx}{\linewidth}{Xccc}
\toprule
Method & Year & DS $\uparrow$ & SR (\%) $\uparrow$ \\
\midrule
UniAD-Base~\cite{Hu_2023_CVPR_UniAD} & 2023 & 45.81 & 16.36 \\
VAD~\cite{Jiang_2023_ICCV_VAD} & 2023 & 42.35 & 15.00 \\
DriveAdapter~\cite{Jia_2023_ICCV_DriveAdapter} & 2023 & 64.22 & 33.08 \\
\midrule
MomAD (SD)~\cite{Song_2025_MomAD} & 2025 & 47.91 & 18.11 \\
WoTE~\cite{Li_2025_WoTE} & 2025 & 61.71 & 31.36 \\
DriveTransformer-L~\cite{Jia_2025_ICLR_DriveTransformer} & 2025 & 63.46 & 35.01 \\
DIVER~\cite{Song_2025_arXiv_DIVER} & 2025 & 68.90 & 36.75 \\
SafeDrive~\cite{Kim_2026_CVPR_SafeDrive} & 2026 & 66.77 & \underline{42.40} \\
Diffusion-Phy~\cite{Hu_2026_PR_DiffusionPhy} & 2026 & 67.49 & 36.51 \\
SoAD~\cite{Gao_2026_TSMC_SoAD} & 2026 & \underline{70.47} & 41.36 \\
\midrule
\rowcolor{gray!12}
FIRE-Drive (Ours) & 2026 & \textbf{71.43} & \textbf{42.73} \\
\bottomrule
\end{tabularx}
\par\vspace{2pt}\raggedright\scriptsize
SD denotes the SparseDrive-based MomAD variant. DIVER uses its highest-DS configuration in the cited paper (arXiv v3, Table I). SoAD includes safety creeping. Training data, sensor and navigation inputs, and auxiliary supervision differ across methods; these are system-level literature comparisons, not matched retrainings.
\end{table}

\subsection{Quantitative Results}

\textbf{NAVSIM.} Table~\ref{tab:navsim_v1_results} compares planning performance under the reported model configurations. Among the listed diffusion-based planners with a ResNet34 backbone, FIRE-Drive achieves the highest PDMS. Relative to DiffusionDrive and DIVER, its results combine stronger progress with higher NC and TTC scores. It also obtains a modest aggregate improvement over DiffusionDriveV2, with particularly strong NC and TTC. The within-framework ablations below examine how initialization, supervision, and scoring contribute to these results.

\textbf{Bench2Drive.} Table~\ref{tab:bench2drive_results} extends the comparison to reactive closed-loop driving. FIRE-Drive achieves higher DS and SR than the listed diffusion-based methods, Diffusion-Phy and DIVER, indicating competitive route-level performance alongside its NAVSIM results. It also exceeds SoAD in both metrics and achieves higher DS with slightly higher SR than SafeDrive. These literature comparisons reflect complete systems under their reported training and evaluation configurations.

\subsection{Ablation Studies}

\begin{table}[!htbp]
\centering
\caption{Progressive component ablation on NAVSIM v1. Higher is better; best results are bolded.}
\label{tab:main_ablation}
\footnotesize
\setlength{\tabcolsep}{2.5pt}
\renewcommand{\arraystretch}{1.08}
\begin{tabularx}{\linewidth}{@{}X c c c c c@{}}
\toprule
Variant & NC & DAC & EP & TTC & PDMS \\
\midrule
DiffusionDrive$^{\dagger}$ & 98.2 & 96.2 & 82.2 & 94.7 & 88.1 \\
\midrule
Feasible anchors & 98.3 & 96.6 & 84.6 & 94.8 & 89.0 \\
+ Candidate supervision & 97.7 & \textbf{97.6} & 87.5 & 93.0 & 90.3 \\
+ Risk-aware scoring & 98.4 & \textbf{97.6} & \textbf{88.4} & 94.6 & 91.3 \\
+ Refinement (Full) & \textbf{98.9} & 97.5 & 87.6 & \textbf{95.8} & \textbf{91.5} \\
\bottomrule
\end{tabularx}
\par\vspace{2pt}\raggedright\scriptsize
$\dagger$ Literature reference. Below the divider, modules are added progressively within our framework. Comfort is 100 throughout and is omitted.
\end{table}

\begin{table}[!htbp]
\centering
\caption{Detailed ablation of key designs on NAVSIM v1. Each variant is obtained by disabling one design from the full model. Best results are bolded.}\label{tab:detailed_ablation}
\footnotesize
\setlength{\tabcolsep}{2.2pt}
\renewcommand{\arraystretch}{1.08}
\begin{tabular*}{\linewidth}{@{\extracolsep{\fill}}p{0.34\linewidth} c c c c c c@{}}
\toprule
Variant & NC & DAC & EP & TTC & C & PDMS \\
\midrule
Full Model & \textbf{98.9} & \textbf{97.5} & \textbf{87.6} & \textbf{95.8} & \textbf{100} & \textbf{91.5} \\
w/o Anchor Constraint & 98.8 & 97.0 & 87.5 & 95.5 & \textbf{100} & 91.1 \\
w/o Mode Risk Context & 98.5 & 97.4 & \textbf{87.6} & 94.2 & 99.8 & 90.7 \\
\bottomrule
\end{tabular*}
\end{table}

\begin{figure*}[!t]
	\centering
	\includegraphics[width=\textwidth]{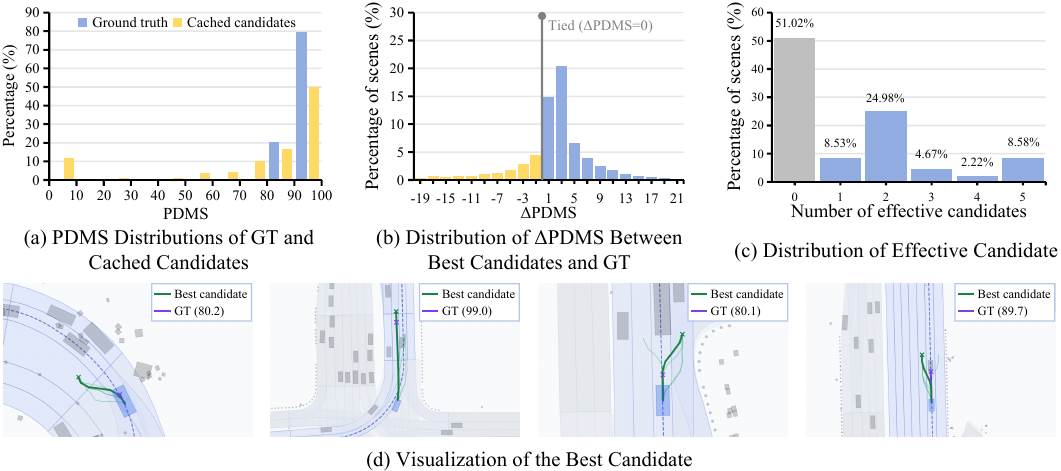}
	\caption{Statistics and visualization of the auxiliary candidate pool. (a) PDMS distributions of ground-truth trajectories and cached candidates, normalized separately by their respective counts. (b) The central range of the per-scene PDMS difference between the best cached candidate and the ground truth; negative tails outside the displayed range are omitted, and ties are shown separately at zero. Percentages use the same training-scene denominator. $\Delta\mathrm{PDMS}=\mathrm{PDMS}_{\mathrm{best\ candidate}}-\mathrm{PDMS}_{\mathrm{GT}}$. (c) The number of effective supervision candidates per scene after validity, geometry, and greater-than-GT filtering. Panels (b) and (c) summarize 103,288 training scenes. (d) Representative cached candidates and ground-truth trajectories.}\label{fig:summ_candi}
\end{figure*}

We conduct all ablations on NAVSIM v1, since it is the primary benchmark of this work. Table~\ref{tab:main_ablation} enables feasible anchor initialization, candidate supervision, risk-aware scoring, and refinement progressively, while Table~\ref{tab:detailed_ablation} evaluates the removal of each of two key designs from the full model.

\textbf{Feasible anchor.} The anchor-enabled variant obtains 89.0 PDMS. The DiffusionDrive score of 88.1 is included as a literature reference. The within-framework removal experiment in Table~\ref{tab:detailed_ablation} evaluates the contribution of the anchor constraint. Figure~\ref{fig:anchor_initialization_comparison} complements this ablation by comparing DiffusionDrive initialization with dynamic anchors before and after feasibility projection. Across the illustrated scenes, the dynamic priors adapt to the local road layout, while projection adjusts the constrained prefix toward the admissible region. These initialization examples are distinct from the final decoded planning trajectories.

\textbf{Candidate supervision.} Compact high-quality candidate supervision further improves PDMS by 1.3 points, especially in EP and DAC. The supervision mask retains auxiliary references that outperform the ground-truth trajectory under the cached evaluator score, providing additional progress-oriented training targets. NC and TTC decrease in this variant, indicating that the progress improvement is accompanied by lower safety-related scores before risk-aware scoring is added. This trade-off clarifies the role of auxiliary supervision: an evaluator-filtered reference set supplies useful alternatives, but does not ensure that the learned decoder reproduces their geometry exactly or that the selector chooses the best resulting mode. Moreover, filtering by an aggregate score does not require improvement in every component metric. The observed pattern therefore motivates coupling broader supervision with candidate-specific risk evaluation.

\begin{figure*}[tp]
	\centering
	\includegraphics[width=\textwidth]{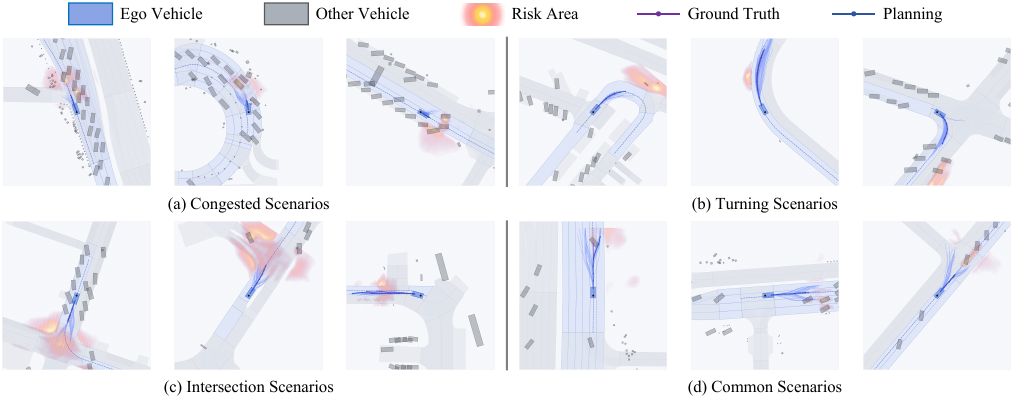}
	\caption{Visualization of risk-aware attention and planned trajectories in twelve examples grouped as (a) Congested Scenarios, (b) Turning Scenarios, (c) Intersection Scenarios, and (d) Common Scenarios. Colored overlays highlight spatial risk evidence around the evaluated trajectories. Purple and blue lines denote the ground-truth and planned trajectories, respectively.}\label{fig:risk_vis}
\end{figure*}

\textbf{Risk-aware scoring.} Adding risk-aware scoring is associated with a PDMS increase from 90.3 to 91.3 in Table~\ref{tab:main_ablation}. NC and TTC recover while EP also increases, suggesting that the added risk evidence helps balance progress and safety within the expanded candidate set. Because the generator and scorer are jointly trained here, this experiment measures their combined response to the added branch. The fixed-input comparison below examines selection more directly.

\textbf{Trajectory refinement.} Refinement provides a modest aggregate improvement, with stronger NC and TTC and slightly lower EP. This pattern is consistent with its role as a local correction of the decoded trajectories.

The detailed ablation in Table~\ref{tab:detailed_ablation} shows that removing the anchor constraint reduces PDMS from 91.5 to 91.1. Here, ``w/o Anchor Constraint'' removes the feasible-region restriction on initialization. The full model achieves modest improvements, most notably in DAC and TTC, while EP changes little and comfort remains unchanged. This pattern supports the use of scene-admissible initialization to improve safety-related performance without reducing progress. The experiment evaluates the semantic--reachability restriction jointly, with the other components present; it does not isolate the two masks or establish an additive gain over the literature baseline.

Removing mode risk context lowers both PDMS and TTC, supporting the use of candidate-conditioned risk evidence in the jointly trained system.

\begin{figure*}[!t]
	\centering
	\includegraphics[width=\textwidth]{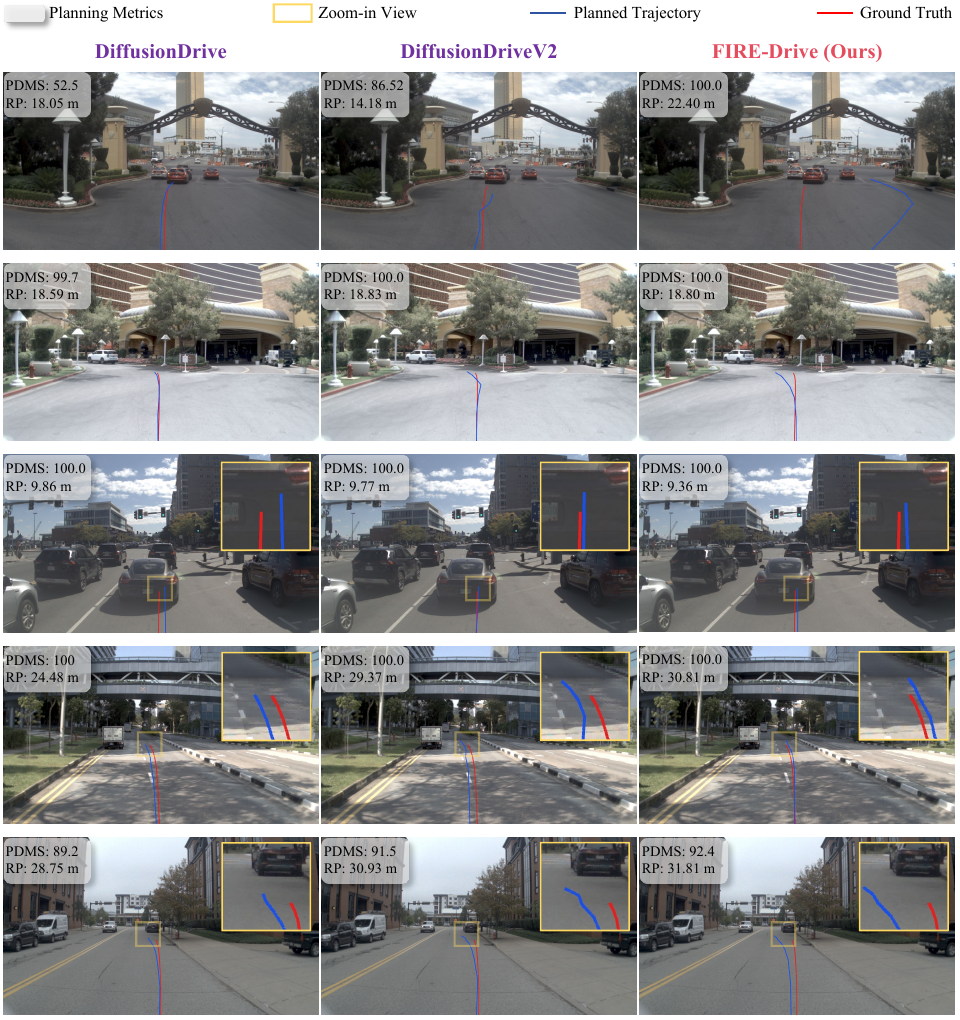}
	\caption{Qualitative comparison of DiffusionDrive, DiffusionDriveV2, and FIRE-Drive on matched scenes. Red and blue lines denote the logged ground-truth trajectory and the planned trajectory, respectively. Yellow boxes mark regions shown in enlarged insets. Gray boxes report PDMS and route progress (RP, in meters).}\label{fig:compare_diff}
\end{figure*}

\subsection{Controlled Ablation of the Trajectory Scorer}

To examine selection under matched inputs, we retrain three scorer variants with the same input features and candidate trajectories and evaluate them on NAVSIM navtest. The full scorer includes mode risk context, semantic risk bias, and risk supervision. The first ablation removes both the risk bias and risk supervision, while the second removes mode risk context. These experiments compare scorer variants under fixed feature and trajectory conditions, separately from the jointly trained system ablations above.

\begin{table}[!htbp]
\centering
\caption{Scorer ablation on NAVSIM navtest with matched input features and candidate trajectories. Scorers are retrained for this experiment. Higher is better for all metrics; best results are bolded.}
\label{tab:scorer_ablation}
\footnotesize
\setlength{\tabcolsep}{3pt}
\renewcommand{\arraystretch}{1.12}
\begin{tabularx}{\linewidth}{Xccccc}
\toprule
Scorer & NC & DAC & EP & TTC & PDMS \\
\midrule
\rowcolor{gray!12}
Full scorer & \textbf{98.4} & \textbf{97.5} & \textbf{88.1} & \textbf{94.1} & \textbf{91.0} \\
w/o Bias and Risk Supervision & 98.0 & 97.1 & 87.9 & 93.8 & 90.4 \\
w/o Risk Context & 98.2 & 97.0 & 87.6 & 94.0 & 90.3 \\
\bottomrule
\end{tabularx}
\end{table}

As shown in Table~\ref{tab:scorer_ablation}, the full scorer improves PDMS by 0.6 points over the variant without risk bias and risk supervision, and by 0.7 points over the baseline scorer~\cite{Guo_2025_arXiv_iPad} without risk context. It also leads across the other reported metrics. With identical candidate trajectories, this comparison isolates the effect of the tested scorer designs on selection and supports the use of explicit risk evidence. The bias-and-supervision ablation evaluates the two components jointly.

The gains across NC, DAC, EP, and TTC suggest that the tested risk-aware scorer improves safety-related scores while also selecting trajectories with slightly greater evaluated progress. This controlled experiment complements the joint-training ablation by holding candidate quality fixed; absolute scores should be compared within each experiment's training conditions.

\subsection{Candidate Statistics and Qualitative Analysis}

Figure~\ref{fig:summ_candi} summarizes the auxiliary reference pool and illustrates representative candidate trajectories. Cached candidates span a range of evaluator scores. During training, the geometry and greater-than-GT masks select the subset used for auxiliary supervision, and scenes with no eligible reference contribute no candidate-supervision loss. The displayed best-candidate statistics describe the reference pool before this training-time selection. Among the 103,288 training scenes, 48.98\% contain at least one effective supervision candidate; the remaining 51.02\% skip candidate supervision.

The variable effective-set size reflects quality-based admission. Strict greater-than-GT filtering excludes ties and cannot admit references when GT already reaches the evaluator ceiling. Other scenes may lack a reference satisfying both geometry and score criteria. Retaining ordinary GT supervision in these cases avoids forcing arbitrary alternatives into training. Thus, the raw score distributions characterize the constructed pool, whereas the post-filter counts quantify its actual contribution to supervision. These statistics support selective auxiliary supervision: the cached pool does not provide a qualifying alternative for every scene. They do not establish that better trajectories are absent outside this pool, but motivate reusing validated references where available.

Figure~\ref{fig:risk_vis} presents three examples in each of four illustrative groups: Congested Scenarios (a), Turning Scenarios (b), Intersection Scenarios (c), and Common Scenarios (d). These groups illustrate dense surrounding traffic, changes in road direction, junction interactions, and routine driving contexts, respectively. The risk overlays illustrate spatial evidence near surrounding vehicles and road boundaries along the evaluated trajectories, showing how candidate-conditioned context varies with the scene. The highlighted risk responses form spatial regions around the evaluated trajectories rather than being confined to a single object, reflecting the surrounding traffic and road-boundary context. These qualitative examples complement the matched-input scorer ablation.

Figure~\ref{fig:compare_diff} compares the three planners on five matched scenes, ordered from top to bottom. In the first scene, FIRE-Drive selects a visibly different lateral alternative and obtains both higher PDMS and greater route progress (RP) than the two baselines. This illustrates the value of selecting a useful alternative rather than merely following the logged trajectory's projected shape. In the fifth scene, FIRE-Drive again achieves the highest PDMS and RP, showing a joint improvement in evaluated driving quality and progress.

The middle scenes reveal more nuanced behavior. In the fourth scene, all methods reach a PDMS of 100, but FIRE-Drive achieves route progress of 30.81\,m, compared with 29.37\,m for DiffusionDriveV2 and 24.48\,m for DiffusionDrive. Its enlarged trajectory segment also follows the logged trajectory's local direction more closely. RP therefore exposes differences that the saturated aggregate score cannot distinguish. In the second scene, FIRE-Drive matches DiffusionDriveV2's maximum PDMS while showing a smoother projected trajectory through the turn. In the third, where a vehicle occupies the lane immediately ahead, it retains maximum PDMS with slightly shorter route progress, consistent with more cautious advancement behind the leading vehicle. Together, these examples show that its gains are scene-dependent rather than arising from uniformly extending trajectories. They complement the quantitative ablations by illustrating both alternative-maneuver selection and local trajectory alignment; RP is interpreted jointly with PDMS and scene geometry.

\clearpage
\section{Conclusions}

This paper presented FIRE-Drive, a diffusion-based E2E planner that improves multimodal planning through feasibility-informed trajectory priors, risk-aware mode evaluation, and compact candidate-guided supervision. The proposed method constrains the initial trajectory prior with short-horizon semantic and reachability information, injects mode-conditioned BEV risk evidence into the scoring branch, and uses high-quality auxiliary candidates with lightweight refinement to strengthen supervision for secondary modes. Experiments on NAVSIM v1 and Bench2Drive show competitive planning performance in non-reactive and reactive evaluation settings. The matched-input scorer ablation further supports the contribution of explicit risk evidence to trajectory selection. The short-horizon initialization constraints leave full-horizon feasibility dependent on subsequent decoding and selection. The current single-frame input also limits temporal modeling, while the auxiliary reference pool covers only a subset of possible behaviors, particularly in highly interactive cases such as being overtaken or emergency avoidance. Future work will focus on temporal modeling and adaptive reference construction for interactive driving.

\FloatBarrier{}

\bibliographystyle{IEEEtran}
\bibliography{references}

@inproceedings{Ho_2020_NeurIPS_DDPM,
  author    = {Jonathan Ho and Ajay Jain and Pieter Abbeel},
  title     = {Denoising Diffusion Probabilistic Models},
  booktitle = {Adv. Neural Inf. Process. Syst.},
  volume    = {33},
  pages     = {6840--6851},
  year      = {2020}
}

@inproceedings{He_2016_CVPR_ResNet,
  author    = {Kaiming He and Xiangyu Zhang and Shaoqing Ren and Jian Sun},
  title     = {Deep Residual Learning for Image Recognition},
  booktitle = {Proc. IEEE Conf. Comput. Vis. Pattern Recognit. (CVPR)},
  pages     = {770--778},
  year      = {2016}
}

@article{Bojarski_2016_arXiv_PilotNet,
  author  = {Mariusz Bojarski and Davide Del Testa and Daniel Dworakowski and Bernhard Firner and Beat Flepp and Prasoon Goyal and Lawrence D. Jackel and Mathew Monfort and Urs Muller and Jiakai Zhang and Xin Zhang and Jake Zhao and Karol Zieba},
  title   = {End to End Learning for Self-Driving Cars},
  journal = {arXiv preprint arXiv:1604.07316},
  year    = {2016}
}

@inproceedings{Dosovitskiy_2017_CoRL_CARLA,
  author    = {Alexey Dosovitskiy and German Ros and Felipe Codevilla and Antonio Lopez and Vladlen Koltun},
  title     = {{CARLA}: An Open Urban Driving Simulator},
  booktitle = {Proc. Conf. Robot Learn. (CoRL)},
  pages     = {1--16},
  year      = {2017}
}

@inproceedings{Codevilla_2018_ICRA_CIL,
  author    = {Felipe Codevilla and Matthias Muller and Antonio Lopez and Vladlen Koltun and Alexey Dosovitskiy},
  title     = {End-to-End Driving via Conditional Imitation Learning},
  booktitle = {Proc. IEEE Int. Conf. Robot. Autom. (ICRA)},
  pages     = {4693--4700},
  year      = {2018},
  doi       = {10.1109/ICRA.2018.8460487}
}

@article{Chitta_2022_TPAMI_TransFuser,
  author  = {Kashyap Chitta and Aditya Prakash and Bernhard Jaeger and Zehao Yu and Katrin Renz and Andreas Geiger},
  title   = {{TransFuser}: Imitation with Transformer-Based Sensor Fusion for Autonomous Driving},
  journal = {IEEE Trans. Pattern Anal. Mach. Intell.},
  volume  = {45},
  number  = {11},
  pages   = {12878--12895},
  year    = {2022},
  doi     = {10.1109/TPAMI.2022.3192765}
}

@inproceedings{Wang_2024_CVPR_DriveWM,
  author    = {Yuqi Wang and Jiawei He and Lue Fan and Hongxin Li and Yuntao Chen and Zhaoxiang Zhang},
  title     = {Driving into the Future: Multiview Visual Forecasting and Planning with World Model for Autonomous Driving},
  booktitle = {Proc. IEEE/CVF Conf. Comput. Vis. Pattern Recognit. (CVPR)},
  pages     = {14749--14759},
  year      = {2024}
}

@article{Liu_2025_TPAMI_HPP,
  author  = {Haochen Liu and Zhiyu Huang and Wenhui Huang and Haohan Yang and Xiaoyu Mo and Chen Lv},
  title   = {Hybrid-Prediction Integrated Planning for Autonomous Driving},
  journal = {IEEE Trans. Pattern Anal. Mach. Intell.},
  year    = {2025},
  doi     = {10.1109/TPAMI.2025.3526936}
}

@inproceedings{Hu_2023_CVPR_UniAD,
  author    = {Yihan Hu and Jiazhi Yang and Li Chen and Keyu Li and Chonghao Sima and Xizhou Zhu and Siqi Chai and Senyao Du and Tianwei Lin and Wenhai Wang and Lewei Lu and Xiaosong Jia and Qiang Liu and Jifeng Dai and Yu Qiao and Hongyang Li},
  title     = {Planning-Oriented Autonomous Driving},
  booktitle = {Proc. IEEE/CVF Conf. Comput. Vis. Pattern Recognit. (CVPR)},
  pages     = {17853--17862},
  year      = {2023}
}

@inproceedings{Jiang_2023_ICCV_VAD,
  author    = {Bo Jiang and Shaoyu Chen and Qing Xu and Bencheng Liao and Jiajie Chen and Helong Zhou and Qian Zhang and Wenyu Liu and Chang Huang and Xinggang Wang},
  title     = {{VAD}: Vectorized Scene Representation for Efficient Autonomous Driving},
  booktitle = {Proc. IEEE/CVF Int. Conf. Comput. Vis. (ICCV)},
  pages     = {8340--8350},
  year      = {2023}
}

@inproceedings{Jia_2023_ICCV_DriveAdapter,
  author    = {Xiaosong Jia and Yulu Gao and Li Chen and Junchi Yan and Patrick Langechuan Liu and Hongyang Li},
  title     = {{DriveAdapter}: Breaking the Coupling Barrier of Perception and Planning in End-to-End Autonomous Driving},
  booktitle = {Proc. IEEE/CVF Int. Conf. Comput. Vis. (ICCV)},
  pages     = {7953--7963},
  year      = {2023}
}

@inproceedings{Jia_2023_CVPR_ThinkTwice,
  author    = {Xiaosong Jia and Penghao Wu and Li Chen and Jiangwei Xie and Conghui He and Junchi Yan and Hongyang Li},
  title     = {Think Twice Before Driving: Towards Scalable Decoders for End-to-End Autonomous Driving},
  booktitle = {Proc. IEEE/CVF Conf. Comput. Vis. Pattern Recognit. (CVPR)},
  pages     = {21983--21994},
  year      = {2023}
}

@inproceedings{Wu_2022_NeurIPS_TCP,
  author    = {Penghao Wu and Xiaosong Jia and Li Chen and Junchi Yan and Hongyang Li and Yu Qiao},
  title     = {Trajectory-Guided Control Prediction for End-to-End Autonomous Driving: A Simple Yet Strong Baseline},
  booktitle = {Adv. Neural Inf. Process. Syst.},
  volume    = {35},
  pages     = {6119--6132},
  year      = {2022}
}

@inproceedings{Jiang_2026_ICLR_VADv2,
  author    = {Bo Jiang and Shaoyu Chen and Hao Gao and Bencheng Liao and Qian Zhang and Wenyu Liu and Xinggang Wang},
  title     = {{VADv2}: End-to-End Vectorized Autonomous Driving via Probabilistic Planning},
  booktitle = {Int. Conf. Learn. Represent.},
  year      = {2026}
}

@inproceedings{Jiang_2023_CVPR_MotionDiffuser,
  author    = {Chiyu {Max} Jiang and Andre Cornman and Cheolho Park and Benjamin Sapp and Yin Zhou and Dragomir Anguelov},
  title     = {{MotionDiffuser}: Controllable Multi-Agent Motion Prediction Using Diffusion},
  booktitle = {Proc. IEEE/CVF Conf. Comput. Vis. Pattern Recognit. (CVPR)},
  pages     = {9644--9653},
  year      = {2023}
}

@inproceedings{Jia_2024_NeurIPS_Bench2Drive,
  author    = {Xiaosong Jia and Zhenjie Yang and Qifeng Li and Zhiyuan Zhang and Junchi Yan},
  title     = {{Bench2Drive}: Towards Multi-Ability Benchmarking of Closed-Loop End-To-End Autonomous Driving},
  booktitle = {NeurIPS Datasets Benchmarks Track},
  year      = {2024}
}

@inproceedings{Dauner_2024_NeurIPS_NAVSIM,
  author    = {Daniel Dauner and Marcel Hallgarten and Tianyu Li and Xinshuo Weng and Zhiyu Huang and Zetong Yang and Hongyang Li and Igor Gilitschenski and Boris Ivanovic and Marco Pavone and Andreas Geiger and Kashyap Chitta},
  title     = {{NAVSIM}: Data-Driven Non-Reactive Autonomous Vehicle Simulation and Benchmarking},
  booktitle = {Adv. Neural Inf. Process. Syst.},
  year      = {2024}
}

@inproceedings{Nayakanti_2023_ICRA_Wayformer,
  author    = {Nigamaa Nayakanti and Rami Al-Rfou and Aurick Zhou and Kratarth Goel and Khaled S. Refaat and Benjamin Sapp},
  title     = {{Wayformer}: Motion Forecasting via Simple and Efficient Attention Networks},
  booktitle = {Proc. IEEE Int. Conf. Robot. Autom. (ICRA)},
  year      = {2023},
  doi       = {10.1109/ICRA48891.2023.10160609}
}

@inproceedings{Shi_2022_NeurIPS_MTR,
  author    = {Shaoshuai Shi and Li Jiang and Dengxin Dai and Bernt Schiele},
  title     = {Motion Transformer with Global Intention Localization and Local Movement Refinement},
  booktitle = {Adv. Neural Inf. Process. Syst.},
  volume    = {35},
  year      = {2022}
}

@inproceedings{Zhou_2023_CVPR_QCNet,
  author    = {Zikang Zhou and Jianping Wang and Yung-Hui Li and Yu-Kai Huang},
  title     = {Query-Centric Trajectory Prediction},
  booktitle = {Proc. IEEE/CVF Conf. Comput. Vis. Pattern Recognit. (CVPR)},
  pages     = {17863--17873},
  year      = {2023}
}

@inproceedings{Liao_2025_CVPR_DiffusionDrive,
  author    = {Bencheng Liao and Shaoyu Chen and Haoran Yin and Bo Jiang and Cheng Wang and Sixu Yan and Xinbang Zhang and Xiangyu Li and Ying Zhang and Qian Zhang and Xinggang Wang},
  title     = {{DiffusionDrive}: Truncated Diffusion Model for End-to-End Autonomous Driving},
  booktitle = {Proc. IEEE/CVF Conf. Comput. Vis. Pattern Recognit. (CVPR)},
  pages     = {12037--12047},
  year      = {2025}
}

@inproceedings{Xing_2025_CVPR_GoalFlow,
  author    = {Zebin Xing and Xingyu Zhang and Yang Hu and Bo Jiang and Tong He and Qian Zhang and Xiaoxiao Long and Wei Yin},
  title     = {{GoalFlow}: Goal-Driven Flow Matching for Multimodal Trajectories Generation in End-to-End Autonomous Driving},
  booktitle = {Proc. IEEE/CVF Conf. Comput. Vis. Pattern Recognit. (CVPR)},
  pages     = {1602--1611},
  year      = {2025}
}

@article{Li_2024_arXiv_HydraMDP,
  author  = {Zhenxin Li and Kailin Li and Shihao Wang and Shiyi Lan and Zhiding Yu and Yishen Ji and Zhiqi Li and Ziyue Zhu and Jan Kautz and Zuxuan Wu and Yu{-}Gang Jiang and Jose M. Alvarez},
  title   = {{Hydra-MDP}: End-to-End Multimodal Planning with Multi-Target Hydra-Distillation},
  journal = {arXiv preprint arXiv:2406.06978},
  year    = {2024},
  doi     = {10.48550/arXiv.2406.06978}
}

@article{Li_2025_arXiv_HydraMDPpp,
  author  = {Kailin Li and Zhenxin Li and Shiyi Lan and Yuan Xie and Zhizhong Zhang and Jiayi Liu and Zuxuan Wu and Zhiding Yu and Jose M. Alvarez},
  title   = {{Hydra-MDP++}: Advancing End-to-End Driving via Expert-Guided Hydra-Distillation},
  journal = {arXiv preprint arXiv:2503.12820},
  year    = {2025},
  doi     = {10.48550/arXiv.2503.12820}
}

@inproceedings{Jia_2025_ICLR_DriveTransformer,
  author    = {Xiaosong Jia and Junqi You and Zhiyuan Zhang and Junchi Yan},
  title     = {{DriveTransformer}: Unified Transformer for Scalable End-to-End Autonomous Driving},
  booktitle = {Int. Conf. Learn. Represent. (ICLR)},
  year      = {2025}
}

@article{Guo_2025_arXiv_iPad,
  author  = {Ke Guo and Haochen Liu and Xiaojun Wu and Jia Pan and Chen Lv},
  title   = {{iPad}: Iterative Proposal-Centric End-to-End Autonomous Driving},
  journal = {arXiv preprint arXiv:2505.15111},
  year    = {2025},
  doi     = {10.48550/arXiv.2505.15111}
}

@article{Yao_2025_arXiv_DriveSuprim,
  author  = {Wenhao Yao and Zhenxin Li and Shiyi Lan and Zi Wang and Xinglong Sun and Jose M. Alvarez and Zuxuan Wu},
  title   = {{DriveSuprim}: Towards Precise Trajectory Selection for End-to-End Planning},
  journal = {Proc. AAAI Conf. Artif. Intell.},
  volume  = {40},
  number  = {14},
  pages   = {11910--11918},
  year    = {2026},
  month   = {Mar.},
  doi     = {10.1609/aaai.v40i14.38178}
}

@article{Sun_2026_arXiv_SparseDriveV2,
  author  = {Wenchao Sun and Xuewu Lin and Keyu Chen and Zixiang Pei and Xiang Li and Yining Shi and Sifa Zheng},
  title   = {{SparseDriveV2}: Scoring is All You Need for End-to-End Autonomous Driving},
  journal = {arXiv preprint arXiv:2603.29163},
  year    = {2026},
  doi     = {10.48550/arXiv.2603.29163}
}

@article{Kirby_2026_arXiv_DrivingOnRegisters,
  author  = {Ellington Kirby and Alexandre Boulch and Yihong Xu and Yuan Yin and Gilles Puy and Eloi Zablocki and Andrei Bursuc and Spyros Gidaris and Renaud Marlet and Florent Bartoccioni and Anh-Quan Cao and Nermin Samet and Tuan-Hung Vu and Matthieu Cord},
  title   = {Driving on Registers},
  journal = {arXiv preprint arXiv:2601.05083},
  year    = {2026},
  doi     = {10.48550/arXiv.2601.05083}
}

@article{Li_2025_arXiv_GTRS,
  author  = {Zhenxin Li and Wenhao Yao and Zi Wang and Xinglong Sun and Joshua Chen and Nadine Chang and Maying Shen and Zuxuan Wu and Shiyi Lan and Jose M. Alvarez},
  title   = {Generalized Trajectory Scoring for End-to-end Multimodal Planning},
  journal = {arXiv preprint arXiv:2506.06664},
  year    = {2025},
  doi     = {10.48550/arXiv.2506.06664}
}

@inproceedings{Liu_2025_arXiv_BridgeDrive,
  author    = {Shu Liu and Wenlin Chen and Weihao Li and Zheng Wang and Lijin Yang and Jianing Huang and Yipin Zhang and Zhongzhan Huang and Ze Cheng and Hao Yang},
  title     = {{BridgeDrive}: Diffusion Bridge Policy for Closed-Loop Trajectory Planning in Autonomous Driving},
  booktitle = {Proc. Int. Conf. Learn. Represent. (ICLR)},
  year      = {2026}
}

@article{Song_2025_arXiv_DIVER,
  author  = {Ziying Song and Lin Liu and Hongyu Pan and Bencheng Liao and Mingzhe Guo and Lei Yang and Yongchang Zhang and Shaoqing Xu and Caiyan Jia and Yadan Luo},
  title   = {{DIVER}: Reinforced Diffusion Breaks Imitation Bottlenecks in End-to-End Autonomous Driving},
  journal = {arXiv preprint arXiv:2507.04049},
  year    = {2025},
  doi     = {10.48550/arXiv.2507.04049}
}

@article{Liu_2026_TPAMI_R2SE,
  author  = {Haochen Liu and Tianyu Li and Haohan Yang and Li Chen and Caojun Wang and Ke Guo and Haochen Tian and Hongchen Li and Hongyang Li and Chen Lv},
  title   = {Reinforced Refinement with Self-Aware Expansion for End-to-End Autonomous Driving},
  journal = {IEEE Trans. Pattern Anal. Mach. Intell.},
  year    = {2026},
  doi     = {10.1109/TPAMI.2026.3653866}
}

@inproceedings{Zheng_2026_CVPR_ResAD,
  author = {Zhiyu Zheng and Shaoyu Chen and Haoran Yin and Xinbang Zhang and Jialv Zou and Xinggang Wang and Qian Zhang and Lefei Zhang},
  title = {{ResAD}: Normalized Residual Trajectory Modeling for End-to-End Autonomous Driving},
  booktitle = {Proc. IEEE/CVF Conf. Comput. Vis. Pattern Recognit.},
  year = {2026},
  pages = {3729--3739}
}

@article{Zou_2025_arXiv_DiffusionDriveV2,
  author  = {Jialv Zou and Shaoyu Chen and Bencheng Liao and Zhiyu Zheng and Yuehao Song and Lefei Zhang and Qian Zhang and Wenyu Liu and Xinggang Wang},
  title   = {{DiffusionDriveV2}: Reinforcement Learning-Constrained Truncated Diffusion Modeling in End-to-End Autonomous Driving},
  journal = {arXiv preprint arXiv:2512.07745},
  year    = {2025},
  doi     = {10.48550/arXiv.2512.07745}
}

@article{Wang_2025_TPAMI_WAKE,
  author  = {Chengyue Wang and Haicheng Liao and Zhenning Li and Chengzhong Xu},
  title   = {{WAKE}: Towards Robust and Physically Feasible Trajectory Prediction for Autonomous Vehicles With WAvelet and KinEmatics Synergy},
  journal = {IEEE Trans. Pattern Anal. Mach. Intell.},
  volume  = {47},
  number  = {4},
  pages   = {3126--3140},
  year    = {2025},
  doi     = {10.1109/TPAMI.2025.3529259}
}

@article{Wang_2024_arXiv_RiskAwareTP,
  author  = {Qingfan Wang and Dongyang Xu and Gaoyuan Kuang and Chen Lv and Shengbo Eben Li and Bingbing Nie},
  title   = {Risk-Aware Vehicle Trajectory Prediction Under Safety-Critical Scenarios},
  journal = {IEEE Trans. Intell. Transp. Syst.},
  volume  = {26},
  number  = {5},
  pages   = {7156--7171},
  year    = {2025},
  doi     = {10.1109/TITS.2025.3525744}
}

@article{Song_2025_MomAD,
  author = {Ziying Song and Caiyan Jia and Lin Liu and Hongyu Pan and Yongchang Zhang and Junming Wang and Xingyu Zhang and Shaoqing Xu and Lei Yang and Yadan Luo},
  title = {Don't Shake the Wheel: Momentum-Aware Planning in End-to-End Autonomous Driving},
  journal = {arXiv preprint arXiv:2503.03125},
  year = {2025}
}

@article{Li_2025_WoTE,
  author = {Yingyan Li and Yuqi Wang and Yang Liu and Jiawei He and Lue Fan and Zhaoxiang Zhang},
  title = {End-to-End Driving with Online Trajectory Evaluation via {BEV} World Model},
  journal = {arXiv preprint arXiv:2504.01941},
  year = {2025}
}

@article{Chen_2024_TPAMI_E2ESurvey,
  author = {Li Chen and Penghao Wu and Kashyap Chitta and Bernhard Jaeger and Andreas Geiger and Hongyang Li},
  title = {End-to-End Autonomous Driving: Challenges and Frontiers},
  journal = {IEEE Trans. Pattern Anal. Mach. Intell.},
  year = {2024}, volume = {46}, number = {12}, pages = {10164--10183},
  doi = {10.1109/TPAMI.2024.3435937}
}

@inproceedings{Shao_2023_CVPR_ReasonNet,
  author = {Hao Shao and Letian Wang and Ruobing Chen and Steven L. Waslander and Hongsheng Li and Yu Liu},
  title = {{ReasonNet}: End-to-End Driving With Temporal and Global Reasoning},
  booktitle = {Proc. IEEE/CVF Conf. Comput. Vis. Pattern Recognit.},
  year = {2023}, pages = {13723--13733}
}

@inproceedings{Sun_2025_ICRA_SparseDrive,
  author = {Wenchao Sun and Xuewu Lin and Yining Shi and Chuang Zhang and Haoran Wu and Sifa Zheng},
  title = {{SparseDrive}: End-to-End Autonomous Driving via Sparse Scene Representation},
  booktitle = {Proc. IEEE Int. Conf. Robot. Autom.},
  year = {2025}, pages = {8795--8801}
}

@article{Huang_2024_TNNLS_DIPP,
  author = {Zhiyu Huang and Haochen Liu and Jingda Wu and Chen Lv},
  title = {Differentiable Integrated Motion Prediction and Planning With Learnable Cost Function for Autonomous Driving},
  journal = {IEEE Trans. Neural Netw. Learn. Syst.},
  year = {2024}, volume = {35}, number = {11}, pages = {15222--15236},
  doi = {10.1109/TNNLS.2023.3283542}
}

@inproceedings{Dauner_2023_CoRL_PDM,
  author = {Daniel Dauner and Marcel Hallgarten and Andreas Geiger and Kashyap Chitta},
  title = {Parting with Misconceptions about Learning-based Vehicle Motion Planning},
  booktitle = {Proc. Conf. Robot Learn.},
  year = {2023}, pages = {1268--1281}
}

@inproceedings{Cheng_2024_ICRA_PlanTF,
  author = {Jie Cheng and Yingbing Chen and Xiaodong Mei and Bowen Yang and Bo Li and Ming Liu},
  title = {Rethinking Imitation-based Planners for Autonomous Driving},
  booktitle = {Proc. IEEE Int. Conf. Robot. Autom.},
  year = {2024}, pages = {14123--14130},
  doi = {10.1109/ICRA57147.2024.10611364}
}

@inproceedings{Song_2021_ICLR_DDIM,
  author = {Jiaming Song and Chenlin Meng and Stefano Ermon},
  title = {Denoising Diffusion Implicit Models},
  booktitle = {Proc. Int. Conf. Learn. Represent.},
  year = {2021}
}

@inproceedings{Janner_2022_ICML_Diffuser,
  author = {Michael Janner and Yilun Du and Joshua Tenenbaum and Sergey Levine},
  title = {Planning with Diffusion for Flexible Behavior Synthesis},
  booktitle = {Proc. Int. Conf. Mach. Learn.},
  year = {2022}, pages = {9902--9915}
}

@inproceedings{Chi_2023_RSS_DiffusionPolicy,
  author = {Cheng Chi and Siyuan Feng and Yilun Du and Zhenjia Xu and Eric Cousineau and Benjamin Burchfiel and Shuran Song},
  title = {Diffusion Policy: Visuomotor Policy Learning via Action Diffusion},
  booktitle = {Proc. Robot.: Sci. Syst.},
  year = {2023}
}

@inproceedings{Zheng_2025_ICLR_DiffusionPlanner,
  author = {Yinan Zheng and Ruiming Liang and Kexin Zheng and Jinliang Zheng and Liyuan Mao and Jianxiong Li and Weihao Gu and Rui Ai and Shengbo Eben Li and Xianyuan Zhan and Jingjing Liu},
  title = {Diffusion-Based Planning for Autonomous Driving with Flexible Guidance},
  booktitle = {Proc. Int. Conf. Learn. Represent.},
  year = {2025}
}

@article{Gao_2026_TSMC_SoAD,
  author = {Yinfeng Gao and Deqing Liu and Yupeng Zheng and Qichao Zhang and Da-Wei Ding and Dongbin Zhao},
  title = {{SoAD}: Safety-Oriented Value Estimation for Enhanced Closed-Loop End-to-End Autonomous Driving},
  journal = {IEEE Trans. Syst., Man, Cybern., Syst.},
  year = {2026}, volume = {56}, number = {8}, pages = {4942--4955},
  doi = {10.1109/TSMC.2026.3688954}
}

@inproceedings{Huang_2023_ICCV_GameFormer,
  author = {Zhiyu Huang and Haochen Liu and Chen Lv},
  title = {{GameFormer}: Game-theoretic Modeling and Learning of Transformer-based Interactive Prediction and Planning for Autonomous Driving},
  booktitle = {Proc. IEEE/CVF Int. Conf. Comput. Vis.},
  year = {2023}, pages = {3903--3913}
}

@inproceedings{Yang_2024_CVPR_DiffusionES,
  author = {Brian Yang and Huangyuan Su and Nikolaos Gkanatsios and Tsung-Wei Ke and Ayush Jain and Jeff Schneider and Katerina Fragkiadaki},
  title = {{Diffusion-ES}: Gradient-free Planning with Diffusion for Autonomous and Instruction-guided Driving},
  booktitle = {Proc. IEEE/CVF Conf. Comput. Vis. Pattern Recognit.},
  year = {2024}, pages = {15342--15353}
}

@inproceedings{Liu_2026_CVPR_GuideFlow,
  author = {Lin Liu and Caiyan Jia and Guanyi Yu and Ziying Song and Junqiao Li and Feiyang Jia and Peiliang Wu and Xiaoshuai Hao and Yadan Luo},
  title = {{GuideFlow}: Constraint-Guided Flow Matching for Planning in End-to-End Autonomous Driving},
  booktitle = {Proc. IEEE/CVF Conf. Comput. Vis. Pattern Recognit.},
  year = {2026}, pages = {3719--3728}
}

@inproceedings{Varadarajan_2022_ICRA_MultiPathPP,
  author = {Balakrishnan Varadarajan and Ahmed Hefny and Avikalp Srivastava and Khaled S. Refaat and Nigamaa Nayakanti and Andre Cornman and Kan Chen and Bertrand Douillard and Chi Pang Lam and Dragomir Anguelov and Benjamin Sapp},
  title = {{MultiPath++}: Efficient Information Fusion and Trajectory Aggregation for Behavior Prediction},
  booktitle = {Proc. IEEE Int. Conf. Robot. Autom.},
  year = {2022}
}

@article{Hu_2026_PR_DiffusionPhy,
  author  = {Jinchao Hu and Zixian Wang and Xiaogang Wu and Yue Hu and Yiqiang Wu and Hao Liu},
  title   = {Kinematic Priors Guide Generative Trajectory Planning for Autonomous Vehicles},
  journal = {Pattern Recognit.},
  year    = {2026},
  doi     = {10.1016/j.patcog.2026.113061}
}

@inproceedings{Kim_2026_CVPR_SafeDrive,
  author = {Kim, Jungho and Oh, Jiyong and Yu, Seunghoon and Shin, Hongjae and Kwak, Donghyuk and Choi, Jun Won},
  title = {{SafeDrive}: Fine-Grained Safety Reasoning for End-to-End Driving in a Sparse World},
  booktitle = {Proceedings of the IEEE/CVF Conference on Computer Vision and Pattern Recognition},
  year = {2026}
}

@article{Xu_2026_TITS_TrainingSurvey,
  author = {Chengkai Xu and Yiming Cui and Jiaqi Liu and Yicheng Guo and Cheng Qin and Geyuan Zhang and Xinwei Dong and Shiyu Fang and Peng Hang and Jian Sun},
  title = {A Survey on End-to-End Autonomous Driving Training From the Perspectives of Data, Strategy, and Platform},
  journal = {IEEE Trans. Intell. Transp. Syst.},
  year = {2026},
  note = {Early access, doi: 10.1109/TITS.2026.3695999},
  doi = {10.1109/TITS.2026.3695999}
}

@article{LeMero_2022_TITS_ImitationSurvey,
  author = {Luc {Le Mero} and Dewei Yi and Mehrdad Dianati and Alexandros Mouzakitis},
  title = {A Survey on Imitation Learning Techniques for End-to-End Autonomous Vehicles},
  journal = {IEEE Trans. Intell. Transp. Syst.},
  volume = {23},
  number = {9},
  pages = {14128--14147},
  year = {2022},
  doi = {10.1109/TITS.2022.3144867}
}

@article{Tian_2025_TITS_RiskField,
  author = {Ye Tian and Huaxin Pei and Yuchen Shi and Yi Zhang and Danya Yao and Li Xiao and Bokui Chen},
  title = {Driving Risk Field Model and Its Application in Trajectory Planning: A New Perspective},
  journal = {IEEE Trans. Intell. Transp. Syst.},
  volume = {26},
  number = {9},
  pages = {13522--13537},
  year = {2025},
  doi = {10.1109/TITS.2025.3569852}
}

@article{Lin_2025_TITS_Contingency,
  author = {Zhihao Lin and Jianglin Lan and Anh-Tu Nguyen and David Flynn},
  title = {Contingency-Aware Spatiotemporal Optimization for Safe Autonomous Vehicle Trajectory Planning},
  journal = {IEEE Trans. Intell. Transp. Syst.},
  volume = {26},
  number = {11},
  pages = {18487--18499},
  year = {2025},
  doi = {10.1109/TITS.2025.3597234}
}

\end{document}